# Comparing Socio-technical Design Principles with Guidelines for Human-centered AI

Thomas Herrmann [0000-0002-9270-4501]

Ruhr-University Bochum, Institut für Arbeitswissenschaft, Bochum, Germany
Thomas.Herrmann@ruhr-uni-bochum.de

**Abstract:** Human-centered AI (HCAI) refers to guidelines or principles that aim on ethically oriented design of systems. We compare HCAI- guidelines with principles of socio-technical systems that emerged in the context of conventional information technology. The comparison leads to a revision of socio-technical heuristics by including aspects of AI-usage. The comparison reveals that continuous evolution is a basic characteristic of socio-technical systems, and that human oversight or interventions and the subsequent appropriation of AI-systems lead to continuous adaptation and re-design of the systems, if autonomy is collaboratively exercised. From a socio-technical point of view, the crucial requirement of transparency has not only to be fulfilled with technical features, but also by contributions of the whole system including human actors. It will be promising for using AI, if not only technical features, but organizational and social practices are socio-technically designed in a way that compensates shortcomings of AI.

## 1 Introduction

A considerable number of different guidelines and principles have been developed in the field of human-centered AI (HCAI) [1] [2]. They focus ethical questions about the effects – such as biases – of AI-outcome on people who use it or about whom decisions are made. Additionally, there is an increasing discussion about the future role of humans who work to accomplish tasks while AI is included. The relation between human work and AI will develop within organizational practices and is a matter of socio-technical design and evaluation. Thus, questions arise of how principles of socio-technical design can be adapted with respect to AI (RQ1) and how a comparison between them and guidelines for HCAI can result into a scientific benefit (RQ2). RQ1 has to take the specific characteristics of AI into account such as “AI offers a higher level of automation and self-direction, requiring less human input” [3, p. 1]. We propose that HCAI should deal with a wide variety of different roles AI can take over such as serving as a tool, running automatized processes, providing decision making, or being a member of hybrid teams [4].

In what follows, we provide background on HCAI-related and socio-technical principles and guidelines (Section 2), we describe the process of the systematic literature research (Section3), present the most relevant aspects of guidelines and principles found within the HCAI discourse (Section 4), provide a revision of socio-technical heuristic (RQ1) (Section 5) and a conclusion by comparing the sociotechnical and the HCAI principles (Section 6).

## 2 Background

One quickly can identify basic contributions that give an overview of ethical principles and guidelines in the context of HCAI. For example, Bingley et al. [5] refer to the European Commission, to national initiatives as well as to contribution of companies and research institutions. They give examples like "… fairness, inclusiveness, reliability, safety, transparency, privacy, security, and accountability …". An appropriate starting point is provided by an European Commission's expert group [6] who require respect for human autonomy, prevention of harm, fairness and explicability, and focus on:

(1) human agency and oversight,
(2) technical robustness and safety,
(3) privacy and data governance,
(4) transparency,
(5) diversity, non-discrimination and fairness,
(6) environmental and societal well-being and
(7) accountability.

A recent paper of an expert group [3] outlines six challenges in the context of applying AI and HCAI: Governance and independent oversight; human well-being, human-AI interaction, responsible design of AI, Privacy, and Design and Evaluation framework. These challenges are not directly elaborated as guidelines but include a wide spectrum of them. Examples are – in the context of human well-being – avoidance of harm, multi-optimization of benefits, agency, trust, accountability, and minimizing frustration, stress, anxiety etc. during human-AI interaction. With respect to privacy, not only secrecy and the protection of one's personality are mentioned but also limitation of reachability. Also relevant are preserving human dignity, safety, and agency. The discussion of governance adds the aspects of integrity and resilience.

Weisz et al. [7] add to the discussion that explainability should be extended by possibilities for exploration. Furthermore, they emphasize variability to make clear that a certain AI-output might only be one variant in the context of several appropriate solutions. Variability should and include multiple outputs and imperfection and become part of users' mental models of AI.

Although the paper on the six challenges values the relevance of the socio-technical perspective, it does not refer to basic socio-technical principles as described for example by Cherns [8], [9] or Mumford [10], and summarized by Clegg [11]. Recently, Herrmann et al. [12] give an updated overview by distilling eight socio-technical evaluation heuristics. They use the term "heuristics" since in the field of rapidly developing IT, principles can only have tentative validity and serve more as rule of thumb. Their heuristics are derived not only from the conventional socio-technical principles, but additionally from five further, closely related fields: human-computer interaction, computer supported cooperative work, process design, privacy, and job design. A total of 17 papers from these six areas were analyzed and 173 items were derived that represent aspects of socio-technical principles. These items were grouped into 13 categories. These categories were contrasted against an empirical data base of 306 socio-technical problems in 13 different cases, such as health care, manufacturing industries, or education. The relevance and external validity of the categories were checked by attempting to assign them to the problems in the data base and to identify unassignable problems, overlaps between the categories, or the necessity for new aspects. As a result, eight heuristics were derived (see Table 3, left column). Only one of the cases that were analyzed by Herrmann et al. [13] refers to problems with using AI. Thus, there is a need to update the proposed socio-technical heuristics with regard to AI.

## 3 Method

To extend the scope of relevant principles and guidelines presented in the background section, we conducted a systematic literature research. Table 1) displays the search terms we have applied and the number of hits as well as the steps of filtering. We used Google Scholar to include a wide scope of interdisciplinary research. From our experience, Google Scholar usually covers what can be found with Web of Science, Scopus, IEEE Explore or the ACM digital library. The search term for the combined result is “Human-centered Artificial Intelligence” AND (“ethical guidelines” OR “ethical principles” OR “design guidelines” OR “design principles”). The search was limited to the years 2014 to 2024 since HCAI is a relatively new field. One could criticize that the focus on " HCAI " represents an inappropriate narrowing of potentially relevant work. However, we realized that at least some authors in the field of HCAI who deal with reviews of guidelines and principles have found relevant papers that were not directly related to HCAI, such as [14] [15] [6] [16] [17].

**Table 3:** Sorting out relevant literature.

| Step | Filtering strategy | Papers |
|---|---|---|
| 1 | “human-centered artificial intelligence” "ethical guidelines" | 219 |
| | “human-centered artificial intelligence” "design principles" | 333 |
| | “human-centered artificial intelligence” "design guidelines" | 163 |
| | “human-centered artificial intelligence” "ethical principles" | 351 |
| 2 | Without duplets | 795 |
| 3 | Of general relevance | 576 |
| 4 | Titles point to an overview over guidelines and principles | 46 |
| 5 | Content contributes to the elaboration of guidelines and principles | 13 |
| 6 | Further papers found by analyzing the 13 papers | 5 |
| | **Number of papers analyzed** | **18** |

To filter the 795 hits (without duplets), we firstly excluded those entries that are not in English, or havw less than 2 citations per year although they were published before 2023. In a further step 3), we removed all those whose title does not suggest that an overview of guidelines or principles is given or which only refer to only single principles or guidelines (like Fairness). We then (step 5) filtered out all entries that use lists of guidelines and principles for specific research questions without expanding or critically reflecting these lists. We also removed entries that refer to a specific domain, take a meta-perspective or focus only on processes for applying guidelines or principles. We carefully analyzed the remaining 13 papers; by checking their reference lists we identified five further relevant papers. We extracted 44 aspects and grouped them into 8 categories to allow for an overview.

## 4 Findings

The results of the grouping procedure are presented with Table 2. An important decision was to find the appropriate level of abstraction. We decided to introduce eight main categories since this level allows a comparison with the socio-technical heuristics. The sub-aspects of each category serve two purposes: On the one hand, they are intended as a reference that shows how the analyzed papers characterize the respective main category. On the other hand, they represent examples for requirements that refer to the specific characteristics of AI.

Transparency and autonomy, in particular human agency, are aspects that are highly relevant in the context of AI but were also discussed since years in the context of automation, e.g. the requirement of keeping the human in the loop. The concept of "human agency" has become particularly important in the context of the new AI capabilities. It emphasizes that humans are not restricted to the role of passive recipients of AI-generated outcomes, but can influence and shape the results of AI as active agents [18, p. 2]. The aspect of fairness is of particular relevance in the context of autonomy, as it is addressed to people who are not directly involved in the socio-technical processes of AI usage, e.g. as users, and are therefore limited to a passive role. Trust could have been an own category. Trustworthiness can be considered as a very general requirement [6]– we assigned it to accountability since we are interested in the features that support trustability – and which similarly accountability requirements.

Tabel 2 also covers contradictory requirements. For example, safety might not be compatible with variance when requiring accuracy and correctness. Variance is offered as a way to handle imperfectness of AI-outcome and to offer the user diverse options between which they can choose on the basis of their own capabilities for decision-making. This example mirrors the general tension between the idea that AI has to be designed and supervised in a way that it provides always appropriate results and the insight that AI might be possibly fallible.

**Table 2:** Eight groups of HCAI-related aspects of guidelines and principles

| **Identified aspects of guidelines and principles** | |
|---|---|
| **Transparency**<br>• Awareness of misuse [14], [19]<br>• Comprehensible AI [20]<br>• Workspace awareness [21]<br>• Transparency and explicability [16], [22], [6]…<br>• Awareness and literacy: [23]<br>• Explainability and exploration [7]<br>• Design for Mental Models, … [7]<br>• Traceability [6] | **Fairness**<br>• Non-discrimination and fairness, e.g. avoidance of unfairhas bias [6] [17], [19] (Bingley et al., 2023<br>• Promotion of human values [17], [19]<br>• Justice, fairness and equity [16], [22]<br>• Dignity [16], [22]<br>• Solidarity [16], [22]<br>• Diversity, [6]<br>• Professional responsibility[17], [19]<br>• Inclusiveness [5] |
| **Autonomy**<br>• Human agency [3]<br>• Human autonomy and oversight, e.g. capability for intervention [6], [15]<br>• Freedom and autonomy [16], [22]<br>• Human-in-the-loop e.g. a stop-button [24]<br>• Human control of technology [17], [19] [7] | **Privacy**<br>• Secrecy [3]<br>• Protection of one's personality [3]<br>• Limitation of reachability [3]<br>• Privacy and data governance (e.g. access to data [6]) |
| **Accountability**<br>• Accountability [6], [25], [3]<br>• Responsibility and accountability [16] [21]<br>• Trust [3] [26] | **Variance**<br>• Variability [7]<br>• Multiple outputs and visualizing differences [7]<br>• Imperfection [7] |
| **Benefits and well-being**<br>• Multi-optimization of benefits [3]<br>• Minimizing frustration, stress, anxiety[3]<br>• Environmental well-being [6]<br>• Sustainable development [25] | **Safety**<br>• Avoidance and prevention of harm [3] [6] [7]<br>• Non-maleficence [16], [22].<br>• Technical robustness and safety [6]<br>• Reliability [5]<br>• Accuracy [24]<br>• Data quality, integrity and access [24] |

Table 3 represents the socio-technical heuristics (left column) and relates them to the aspects of AI principles and guidelines (right column) which are relevant for the AI-related revision of these heuristics.

## 5 AI-related revision of sociotechnical heuristics

In what follows we use the findings of Section 4 to revise the socio-technical heuristics as they have been documented more explicitly with the English version of https://hi4.iaw.rub.de/#!/manual (retrieved on 02/04/2024). We include the AI-related aspects in the description of the heuristics and mark them in italics.

**Table 3:** Extending socio-technical heuristics by considering AI

| No. | **Socio-technical heuristics according to Herrmann** et al. [12] | **Related aspect of AI-related principles and guidelines** |
|---|---|---|
| 1 | Visibility about task handling and feedback about its success | Transparency, Accountability, Privacy |
| 2 | Flexibility for variable task handling leading to … evolution of the system | Autonomy, Variance |
| 3 | Communication support for task handling and social interaction | Benefits and wellbeing, accountability, selected aspects of privacy: limited reachability, |
| 4 | Purpose-orientated information exchange for facilitating mental work | Privacy and data governance, selected aspects of safety: data quality, integrity and access; |
| 5 | Balance of effort and experienced benefit by organizational structuring of tasks | Benefits and wellbeing, selected aspects of fairness, such as promotion of human values |
| 6 | Compatibility between requirements, development of competencies, and the system's features | Selected aspects of fairness: avoidance of biases |
| 7 | Efficiency-oriented allocation of tasks for pursuing holistic goals | Benefits and wellbeing |
| 8 | Supportive technology and resources for productive and flawless work | Safety |

**1) Visibility about task handling and feedback about its success**

The status and progress of workflows and technical procedures *into which AI is integrated* are visible and actively explorable as far as it is relevant for task processing and permissible from a privacy point of view. *AI can help to support this visibility. AI-outcome is also a subject of explainability and explorability where users can experimentally research the behavior of AI, e.g. by interventions. Not only the AI-system itself but also other human actors should contribute to explaining the behavior of AI*. Visibility also includes explainability of the background of the socio-technical system, so that one understands why certain events or outcome occur or not, whether certain effects can be expected or not, and how they affect people within and outside the socio-technical system – *also with respect to fairness*. *This includes an understanding of the data basis being used for the training of machine learning components.* Visibility also applies to possible further collaborative work steps and *to the options for individually and collaboratively adapting AI-outcome,* and for further developing and adapting the system*, including AI components.* One can see what one is contributing and what others are contributing, *and – with respect to accountability – which of others' contributions result from employing AI.* The representation of the information for the purpose of visibility must be well understandable *and comprehensible* [20]. Accordingly, one can individually and purposefully select and adapt the extent as well as the degree of abstraction of this information. *This adaptation and personalization of visibility can be supported by AI.* Awareness for the behavior of other agents – *human as well as AI* – serves as a basis for giving appropriate feedback on collaborative work. *Understanding the background includes knowledge about which interests have been involved in the design, training and selection of a system that might include AI-components. Thus, possible interests in misuse or in causing bias are detectable.*

Regular and timely feedback helps to understand how far one has met the expectations of others and *what has been contributed in addition to AI-outcome. Feedback and Visibility have to serve as a prerequisite for AI-related trust building.*

**2) Flexibility for variable task handling leading to a participatory evolution of the system**

One can vary manifold options of task handling and can flexibly decide about technology usage, time management, sharing of tasks etc. *This includes autonomy for teams of human actors who possibly control AI-based processes – such as autonomous driving – and decide whether or how an AI-outcome is included in subsequent workflows.* Consequently, human actors can develop a wide range of competences that support their participation in the ongoing evolution of the whole socio-technical system.

Flexibility, freedom of decision and room for a broad scope of actions open the way to the evolution of the overall system, *e.g. by interactive machine learning* [27]. By minimizing strict rules for how to run tasks, different ways of *decision making* [28] and task handling become possible; this concerns methods, tools, the exchange of information, time management, sequences of action, etc. *For example, people should be able to choose between different AI-systems to be included and they should be able to decide when in the steps of task handling AI is employed* [29]. Groups can flexibly share tasks among themselves *particularly whether an AI-agent is included or not.* The way of using technology can also be varied and includes its adaptability. Workload and stress – *e.g. through the need for oversight of complex AI-processes* – can be mitigated by having different approaches on a team's disposal. *Possibilities for intervention [30] have to be provided that allows for interruptions and phases of fine-grained control without terminating or manipulating all facets of an AI-process.*

By exploiting flexibility, competences are simultaneously developed in a holistic way that promotes participation in the further development of the system. *Consequently, AI has to offer modes of usage that promots the development of human competencies and capabilities* [13]. The coupling of flexibility with participation in this further development is realized in such a way that one can react to systemic interactions, *imperfection* [7]*,* incompleteness, contingency, social dynamics and contextual changes such as renewal of technologies. This includes the development of people's personality to enable them to process tasks more efficiently or to take on new tasks, *or to specify new ways of task sharing between humans and AI.*

*With respect to employing AI-generated outputs, users as well as indirectly affected members of the socio-technical system should not remain passive recipients* [18] *but active participants who can influence and shape the outcomes of AI interactions or even veto them* [31]. *Offering a variance of different outcomes between which the participants can choose or that can be adapted, contributes to flexible usage of AI. Interactive usage of AI allows for exploring the capacities and reliability of a systems and thus helps to build trust* [32] *in the context of possible imperfectness and continuous evolution.*

**3) Communication support for task handling and social interaction**

By technical and spatial support for communication one can be reached for purposes of task handling and coordination. Furthermore, this support is inevitably intertwined with building social relations that include negotiating the duties and rights of roles, or conversation about values, so that reciprocal reliability can be developed. *Opportunities for communication between people is an aspect of well-being that must not be disturbed by AI that mimics human beings.* Opportunities for communication can be established by organizational practices, technical media and spatial arrangements. *AI can help to find communication partners and channels on the one hand and can be used to replace human communicators, e.g. with chatbots, on the other hand. As a prerequisite of accountability, the difference between both options has to be crystal clear. The role an AI-agent might take over as a teammate must be negotiable by communicational means.*

Informal communication has to be maintained and promoted. It is less task-related but more relevant for building social relations and it contributes to the bridging of hierarchies, preservation of confidentiality and trust building. *Informal communication is relevant for trust calibration in the case of employing AI, since trust arises in network relationships (e.g., A trusts an AI system, if A trusts B and B trusts the system).* The extent of reachability is relevant for people's privacy and must be controllable in order to avoid interruptions of communication and task handling. *AI can serve as a gatekeeper to help regulate reachability. Integrating AI into teams and into organizational practices requires new social relations between humans and between humans and AI* [33]. These relationships have to be negotiated, established and maintained via communication between the involved stakeholders.

**4) Information exchange for facilitating mental work**

To support task handling, information is purposefully exchanged via technical means, updated, kept available and minimized. This implies that information items are technically linked with each other, and new information can be derived that possibly violates privacy rights, e.g. in the case of personal profiles. *AI can help to trace the origin of information, and the information from which AI-generated output is derived has also to be traceable. At least, AI has to provide documents and information that back its outcome.*

In order to complete tasks, the necessary information is made available systematically, comprehensibly and situation-related (right time, right place) with the help of technology. *AI can be employed to analyze contextual clues that indicate when which information should be provided.* Thus, no one has to memorize data or struggle with information over-load. People have access to the data they need or have created. The quality, security and accountability of information is technically supported, e.g. by regular archiving, updating and deletion; the conversion of data for the purpose of transfer between different media types is avoided. *Documentation activities and increasing data quality can be supported and automated by AI.*

The availability of information must not violate privacy or confidentiality. *Personal data must not be hidden in large language models or foundation models* [34]. *AI has to be employed to detect and eliminate personal data, e.g. in pictures made in the public.* For those affected, the processing of their personal data or their virtual image is transparent and traceable. Minimization of data and of their accessibility ensures privacy and protects trust, as does self-determination about the information that is transmitted, processed and linked, *or used for the training of machine learning systems*.

**5) Balance between effort and experienced benefit**

Tasks are assigned to people, pooled, and technically supported in a way that make sense and provide fun for people. Thus, a sustaining balancing of efforts and personal benefits is pursued by organizational practices and technical artefacts *including AI-components.*

The handling of tasks is pooled in appropriate task bundles that are meaningful for the work force and is distributed in such a way between persons and technical support that a balanced relationship between benefit and effort can be experienced. *This applies particularly to the task sharing between humans and AI and to the integration of AI-teammates. In accordance with the concept of hybrid intelligence systems [2], the strengths of humans and of AI have to be optimally combined, not only to achieve the best possible results but also to improve beneficial experience during work.* Motivation and sense of fun are also promoted by the fact that the challenges of the task accomplishment correspond to the individual mental, physical and social abilities. The degree of beneficial experience might be technically measured – *possibly with AI-components* – to provide feedback to those who are affected or responsible. Individual preferences, goals, values, and interests are taken into account to achieve the balance; health impairments and unsolicited stress are avoided. *AI helps to reduce stress by offering the interaction-free usage of automated processes or by taking over tasks of routinized documentation. However, the need for oversight and being in control can cause additional effort and stress that have to balanced.* The complete competence spectrum of a person or team and their different communication needs are considered. The balance must also be experienced at the level of groups and organizations. Effort and benefit are not only balanced in everyday work, but also when employees participate in the further development of the system. *The effort of exercising autonomy when dealing with AI must be balanced by the experienceable benefits; this applies also when people participate in the continuous evolution of AI.*

**6) Compatibility between requirements, development of competences and the system's features**

Technical and organizational features of the system are continuously adjusted to each other. They have to meet – within in clarified limits – the requirements from outside in a way that is based on the development of competencies and proactive help for dealing with changing challenges. People's tasks must comply with their technical, social and physical competences and skills. *This is specifically true for AI-related requirements such as using transparency, exercising autonomy, providing accountability, or ensuring fairness. Not only technical, but also organizational practices must support people in meeting these requirements.*

Through continuous adaptation of the socio-technical system and continuous individual and organizational learning, the characteristics of the system fit into the direct organizational environment and meet the requirements imposed on an organization from outside. This fit concerns the language used, legal and ethical aspects – *such as fairness and avoidance of bias in the case of AI –*, social dynamics, goals, processes, physical and technical conditions, etc.

The request to achieve compatibility in order to fulfill external requirements can only be fulfilled within reasonable limits, which must be clarified and comprehensible. In order to achieve compatibility with external requests, corresponding internal compatibility must be maintained: The various components of the overall socio-technical system must support each other consistently and expectably. *For example, AI must transparently reflect the values and goals of the actors involved, e.g. for decision-making or the execution of autonomous processes.* This includes, for example, mutual assistance, instructions and hints (prompting) as a contribution to holistic skills development. The skills are required in order to be prepared for upcoming and future tasks as well as regular changes in conditions and challenges. *According to the concept of hybrid intelligence, the reciprocal, interactive learning of both sides, AI and humans, must contribute to the necessary compatibility of skills.*

**7) Efficiency-oriented allocation of tasks for pursuing holistic goals**

By appropriate sequencing, integration and distribution of tasks – between humans and technology – seamless collaboration is supported. Unnecessary steps or waste of resources are avoided. If needed, an increase of efficiency can be realized.

People are supported so that they can efficiently manage their tasks and workflows without obstacles, health risks, etc. This is achieved through an efficient organization of workflows, for example through a suitable sequence, division or grouping of tasks. *AI can help – e.g. via process monitoring – to optimize workflows that include human work.* Tasks are bundled in such a way that the achievement of holistic goals can be experienced. *AI must not be integrated into workflows in a way that prevents people from understanding the key objectives to be achieved.* Unnecessary tasks or inflexible processes must not be enforced. Waste of resources and the involvement of unnecessary persons or departments are avoided. *AI can be employed to optimize the use of resources.* This includes assistance and quality controls to avoid mistakes or their consequences. For example, tasks are not continued if intermediate results are faulty and unusable. Thus, resources are used sparingly, and tasks are shared between people and distributed between people and technology – *e.g. AI-agents* – in such a way that efficiency is achieved. An increase in performance is made possible by the continuous further development of organization and technology, *e.g. retraining of machine learning components. When comparing the efficiency between conventional technology and AI, the whole AI-lifecycle has to be taken into account, including design, training, organizational integration and retraining as well as the effort of exercising control.*

**8) Supportive technology and resources for productive and flawless work**

Technology and further resources support work and collaboration and consider the intertwining of criteria such as technology acceptance, usability and accessibility for different users, avoiding consequences of mistakes or of misuse, security, and constant updating.

Technology and additional resources are available at the right time and are being further developed according to current possibilities so that task processing is simple and robust against errors. Accordingly, access to the resources is uncomplicated and reliable; they are easy and quick to use (usability) and allow to accelerate the use with increasing experience. Individual human limitations are taken into account and acceptance barriers are reduced. *AI can help to optimize the adaptability of human-computer interfaces.*

The loss or non-availability of data as well as unnecessary waiting times are example of most striking problems to be avoided. Reliability and robustness prevent individually or jointly caused errors as well as unintentional misconduct and prevent deliberate misuse. The entire system can quickly return to an error-free state or revise unwanted effects. *For AI-interfaces, it is important that automated processes can be temporarily interrupted, e.g. to allow users to take over control. Interactive tools are needed to deal with imperfect results* [32]. *User-driven possibilities for retraining must be offered* [27], *and within training material, the data included that has caused erroneous results must be detectable.*

## 6 Discussion and Conclusion

The question (RQ1) of how principles of socio-technical design can be adapted with respect to AI is answered by Section 5. The second research question (RQ2) asks how a comparison between them and guidelines for HCAI can result into a scientific benefit.

There are clear overlaps between socio-technical heuristics for evaluating conventional information technology and for AI usage. The relation between visibility and transparency (see Table 3) is a highly relevant example. In the study of Herrmann et al. [13], visibility proved to be the most important recommendation since it got the highest number of assignments (>20%). The black box character of AI increases the problems of invisibility. From a socio-technical point of view, visibility has not only to be provided for the direct users but also for indirectly affected people such as patients in health care [35], [36]. Visibility or transparency have to take privacy concerns into account and are a crucial prerequisite for trustability and trust calibration [37]. This influence of visibility has been given greater emphasis with the emergence of HCAI. AI can not only decrease but also increase the possibilities for visibility by providing explainability and personalization of explanations by taking mental models into account. However, from a socio-technical point of view, visibility is a requirement that has not only to be fulfilled with technical features but also by contributions of the whole system including human actors. If users do not comprehend the output of an AI-system, they also must have the possibility to ask human experts, such as data analysts, AI-developers or domain experts, for explanations. This example points towards a general recommendation that is neglected in the HCAI discourse: It will be promising for using AI if not only technical features, but organizational and social practices are socio-technically designed in a way that compensates shortcomings of AI.

Similarly, promoting autonomy and human control that allows for flexible, autonomous task handling and AI usage is a requirement that has to be addressed by the whole socio-technical system. It is not sufficient to address this requirement with interface design and functionality of AI. By contrast, organizational practices that promote humans capabilities and readiness for collaboratively staying in control are most crucial to guarantee autonomy [38]. Comparing the socio-technical approaches with HCAI regarding autonomy or flexibility reveals further insights:

- The relation between control, interactive usage and exploration on the one hand and trust building on the other hand is not sufficiently present in the discussion on socio-technical systems and should be given more attention.
- Completely stopping an autonomous process or decision-making workflows that include AI can be risky, especially if not everyone involved is aware of the stop [39]. Temporary interventions are more appropriate as they allow for limited interruptions or have restricted effects, support a "what-if" exploration, and can be easily revised or canceled.
- Continuous evolution is a basic characteristic of socio-technical systems. Human oversight and interventions and the subsequent appropriation of AI-systems [40] lead to continuous adaptation and re-design of the systems when autonomy is collaboratively exercised. This interrelationship should be taken into account more rigorously in the context of HCAI.

The heuristics of Herrmann et al. [12] differentiate between communication and information exchange to emphasize that informal communication is crucial to support building of social relationships as well as the negotiation of rights and duties. Since trust building is embedded in social relationships, a sufficient amount of human communication is necessary, in which the communicators cannot be replaced by AI agents. Furthermore, the difference between the heuristic 'balance of effort and experienced benefit' and the heuristic 'efficiency-oriented allocation of tasks' is important for the HCAI discussion: On the one hand, people may be willing to exercise control and oversight – even if this causes inefficiencies – as long as the invested efforts supports the feeling of being in control and allows for freedom of decision. Autonomy and exercising control can be a value for its own. On the other hand, efficiency is an important aspect where AI can help to reduce the waste of resources and lead to economic benefits.

Methods and criteria that support the design and evaluation of socio-technical systems are widely focused on the interests and wellbeing of those people who are part of the system. The aspect of fairness is therefore underestimated in conventional socio-technical discourses insofar as it addresses those being affected outside the socio-technical system. Only if fairness is reflected in the values of the actors within the socio-technical system and becomes part of their responsibility will it be sufficiently taken into account. Consequently, an additional socio-technical principle or heuristic such as 'value implantation# could be relevant.

Concludingly, HCAI might expand research of how social and organizational practices contribute to mitigate AI-related problems. A continuous evolution of socio-technical systems that include AI has to be promoted as a means to deal with the general imperfectness and fallibility of AI. For research that pursues the socio-technical perspective in HCAI, the relevance of fairness and accountability, particularly of traceability has to be given more attention. It has to be realized that socio-technical requirements are not only meant to regulate AI or to inform its design and usage, but also that AI can help to meet these requirements as a valuable component within socio-technical processes.